\documentclass[pretty, 12pt]{IEEEtran}
\usepackage{mathtools}
\usepackage{graphics}
\usepackage{graphicx}
\usepackage{epstopdf}
\usepackage{lipsum}
\usepackage{float}
\usepackage{textcomp}
\usepackage{adjustbox}
\usepackage{tikz}
\usepackage[section]{placeins}
\usepackage{subcaption}
\usepackage[hang,flushmargin]{footmisc}

\definecolor{yellow}{RGB}{255,255,0}
\definecolor{red}{RGB}{255,91,51}
\definecolor{blue}{RGB}{8,191,223}
\definecolor{green}{RGB}{54,203,30}
\definecolor{grey}{RGB}{170,170,170}
\definecolor{black}{RGB}{0,0,0}
\definecolor{orange}{RGB}{255,140,0}
\definecolor{yellow}{RGB}{225,249,27}

\def\footnoterule{\relax%
  \kern15pt
  \hbox to \columnwidth{\hfill\vrule width 1\columnwidth height 0.6pt\hfill}
  \kern4.6pt}

\DeclareMathOperator*{\argmin}{argmin}

\begin{document}

\tikzset{VertexStyle/.style = {shape = rectangle,fill = gray}}

\title{ Robust Navigation In GNSS Degraded Environment Using Graph Optimization \\ }
\author{Ryan M. Watson and Jason N. Gross , { \it West Virginia University} }
\maketitle

\begin{abstract}
 Robust navigation in urban environments has received a considerable amount of both academic and commercial interest over recent years. This is primarily due to large commercial organizations such as Google and Uber stepping into the autonomous navigation market. Most of this research has shied away from Global Navigation Satellite System (GNSS) based navigation. The aversion to utilizing GNSS data is due to the degraded nature of the data in urban environment (e.g., multipath, poor satellite visibility). The degradation of the GNSS data in urban environments makes it such that traditional (GNSS) positioning methods ( e.g., extended Kalman filter, particle filters) perform poorly. However, recent advances in robust graph theoretic based sensor fusion methods, primarily applied to Simultaneous Localization and Mapping (SLAM) based robotic applications, can also be applied to GNSS data processing. This paper will utilize one such method known as the factor graph in conjunction several robust optimization techniques to evaluate their applicability to robust GNSS data processing. The goals of this study are two-fold. First, for GNSS applications, we will experimentally evaluate the effectiveness of robust optimization techniques within a graph theoretic estimation framework. Second, by releasing the software developed \footnote{The software developed for this study can be obtained from the following link: \href{https://github.com/wvu-navLab/RobustGNSS}{https://github.com/wvu-navLab/RobustGNSS}} and data sets used for this study, we will introduce a new open-source front-end to the Georgia Tech Smoothing and Mapping (GTSAM) library for the purpose of integrating GNSS pseudorange observations.
\end{abstract}

\section*{Introduction}
Traditionally, GNSS data is not utilized to its full potential for autonomously navigating vehicles in urban environments. This is largely ascribable to the possibility of GNSS observables be degraded (e.g., multipath, poor satellite geometry) when confronted with an urban environment. However, the inclusion of GNSS data in such systems would provide substantial information to their navigation algorithms. So, the ability to safely incorporate GNSS observables into existing inference algorithms, even when confronted with environments that have the potential to degrade GNSS observables, is of obvious importance.

To overcome the aforementioned issues, we will leverage the advances made within the robotics community surrounding graph-based simultaneous localization and mapping (SLAM) to efficiently and robustly process GNSS data. Within this community, the advances surrounding optimization have been made on two fronts: optimization efficiency and robustness. For robust graph optimization, the literature can be divided into two broad subsets: traditional M-Estimators~\cite{robustStats}, and more recent graph based robust methods~\cite{switchConstraints, DCS, maxmix}. This paper will evaluate the effectiveness of the mentioned robust optimization techniques when applied specifically to GNSS pseudorange data processing.

The remainder of this paper is organized as follows. First, the technical approach utilized for this study is described, which begins with a discussion on factor graph optimization and then evolves into a discussion on how to make that optimization more robust to erroneous data. Next, the experiential setup and collected data sets are discussed. Then, the results obtained using the previously described models and data sets are provided. Finally, the paper ends with concluding remarks and proposed steps for continued research.

\section{Technical Approach}

This section provides concise overview of the sensor fusion approach utilized in this study. For the reasons mentioned above, it was decided to use a graph-theoretic approach for sensor fusion as opposed to the more common Kalman~\cite{kalman} or particle filters~\cite{particleFilters}. To describe our approach, first, the factor graph~\cite{factorGraphs} is discussed. Then, a discussion is provided on the incorporation of GNSS pseudorange observables into the factor graph framework. Finally, a discussion is provided on methods to make these graph theoretic approaches more tolerant to measurement faults, specifically, in the context of GNSS.

\subsection{Factor Graphs}

The factor graph was purposed in ~\cite{factorGraphs} as a mathematical tool to factorize functions of many variables into smaller subsets. This formulation can be utilized to infer the posterior distribution of the GNSS based state estimation problem. The factorization is represents as a bipartite graph, $\mathcal{G}= (\mathcal{F},\mathcal{X},\mathcal{E})$, where there are two types of vertices: the states to be estimated, $\mathcal{X}$, and the probabilistic constraints applied to those states, $\mathcal{F}$. An edge $\mathcal{E}$ only exists between a state vertex and a factor vertex if that factor is a constraint applied on that time-step. An example factor graph is depicted in Figure 1, where $X_n$ represents the states to be estimate at time-step $n$ (i.e., user position, user velocity, zenith point troposphere delay, GNSS receiver clock bias), $e$ represents the constraint applied to the state by a measurement (i.e., a GNSS pseudorange observable), and $b_n$ represents a probabilistic constraint applied between time-steps (e.g., incorporation of inertial navigation into the factor graph).

\begin{figure}[H]
 \begin{center}
  \begin{adjustbox}{width=0.5\textwidth}
   \begin{tikzpicture}
    \node[shape=circle,fill=grey, minimum size=1.5cm] (A) at (0,0) {$X_{n-2}$};
    \node[shape=circle,fill=grey, minimum size=1.5cm] (B) at (5,0) {$X_{n-1}$};
    \node[shape=circle,fill=grey, minimum size=1.5cm] (C) at (10,0) {$X_{n}$};

    \node[shape=rectangle,fill=blue, minimum size=0.75cm] (D) at (-2,2) {$e_{1}$};
    \node[shape=rectangle,fill=blue, minimum size=0.75cm] (E) at (-1.5,3) {$e_{2}$};
    \node[shape=circle,fill=black,minimum size=0.1cm] (a) at(-0.75,3.5) {};
    \node[shape=circle,fill=black,minimum size=0.1cm] (b) at(-0,3.75) {};
    \node[shape=circle,fill=black,minimum size=0.1cm] (c) at(0.75,3.5) {};
    \node[shape=rectangle,fill=blue, minimum size=0.75cm] (F) at (1.5,3) {$e_{m-1}$} ;
    \node[shape=rectangle,fill=blue, minimum size=0.75cm] (G) at (2,2) {$e_{m}$};

    \node[shape=rectangle,fill=green, minimum size=0.75cm] (P) at (2.5,0) {$b_{n-1}$};

    \node[shape=rectangle,fill=blue, minimum size=0.75cm] (H) at (3,2) {$e_{1}$};
    \node[shape=rectangle,fill=blue, minimum size=0.75cm] (I) at (3.5,3) {$e_{2}$};
    \node[shape=circle,fill=black,minimum size=0.1cm] (a) at(4.25,3.5) {};
    \node[shape=circle,fill=black,minimum size=0.1cm] (b) at(5,3.75) {};
    \node[shape=circle,fill=black,minimum size=0.1cm] (c) at(5.75,3.5) {};
    \node[shape=rectangle,fill=blue, minimum size=0.75cm] (J) at (6.5,3) {$e_{m-1}$} ;
    \node[shape=rectangle,fill=blue, minimum size=0.75cm] (K) at (7,2) {$e_{m}$};

    \node[shape=rectangle,fill=green, minimum size=0.75cm] (Q) at (7.5,0) {$b_n$};

    \node[shape=rectangle,fill=blue, minimum size=0.75cm] (L) at (8,2) {$e_{1}$};
    \node[shape=rectangle,fill=blue, minimum size=0.75cm] (M) at (8.5,3) {$e_{2}$};
    \node[shape=circle,fill=black,minimum size=0.1cm] (a) at(9.25,3.5) {};
    \node[shape=circle,fill=black,minimum size=0.1cm] (b) at(10,3.75) {};
    \node[shape=circle,fill=black,minimum size=0.1cm] (c) at(10.75,3.5) {};
    \node[shape=rectangle,fill=blue, minimum size=0.75cm] (N) at (11.5,3) {$e_{m-1}$} ;
    \node[shape=rectangle,fill=blue, minimum size=0.75cm] (O) at (12,2) {$e_{m}$};

    \path [-,line width=1pt] (A) edge node {} (D);
    \path [-,line width=1pt] (A) edge node {} (E);
    \path [-,line width=1pt] (A) edge node {} (F);
    \path [-,line width=1pt] (A) edge node {} (G);

    \path [-,line width=1pt] (A) edge node {} (P);
    \path [-,line width=1pt] (P) edge node {} (B);

    \path [-,line width=1pt] (B) edge node {} (H);
    \path [-,line width=1pt] (B) edge node {} (I);
    \path [-,line width=1pt] (B) edge node {} (J);
    \path [-,line width=1pt] (B) edge node {} (K);

    \path [-,line width=1pt] (B) edge node {} (Q);
    \path [-,line width=1pt] (Q) edge node {} (C);

    \path [-,line width=1pt] (C) edge node {} (L);
    \path [-,line width=1pt] (C) edge node {} (M);
    \path [-,line width=1pt] (C) edge node {} (N);
    \path [-,line width=1pt] (C) edge node {} (O);

   \end{tikzpicture}
     \end{adjustbox}
   \end{center}
  \caption{Example factor graph for GNSS data processing}
  \end{figure}
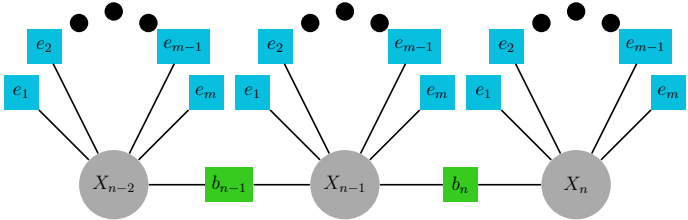

  In the graph, $e_i$ represents an error function or probabilistic constraints applied to the state at the specified time-step. When Gaussian noise is assumed, the error function is equivalent innovation residual of the traditional Kalman filter, as shown in Eq 1. Utilizing this information it is easy to see that the optimal state estimate can be calculated through a traditional non-linear least squares formulation (i.e., Levenberg Marquardt ~\cite{LM}) that minimizes the error over the graph.

  \begin{equation}
   e_i = H_i(X_i) - Z_i
  \end{equation}

  \begin{equation}
   \hat{X} = argmin \sum_i  \lvert \lvert e_i \rvert \rvert^{2}_{\Sigma}
  \end{equation}



  \subsubsection{Pseudorange Factor}
  With the general factor graph framework discussed, the discussion can now move to optimization of GNSS navigation applications. To construct the pseudorange factor, the dual-undifferenced GNSS observables are utilized. Because undifferenced data is being used, methods for mitigating other GNSS error-sources (e.g., troposphere and ionosphere delays, and GNSS receiver clock bias) are incorporated in the measurement models.

  To mitigate the ionospheric delay, the dispersive nature of the ionosphere is leveraged, and a linear combination of the GPS $L_{1}$ and $L_{2}$ frequencies (1575.42 MHz and 1227.60 MHz, respectively) is formed to produce ionospheric-free (IF) pseudorange measurements in order to eliminate the delay to first order \cite{misra2006global}. The IF pseudorange combination can be seen in Eq. \ref{PR}: where, $f_{1}$ and $f_{2}$ are the $L_{1}$ and $L_{2}$ frequencies, and $\rho_{L1}$ and $\rho_{L2}$ are the pseudorange measurements on the $L_{1}$ and $L_{2}$ frequencies. The superscript $j$ in Eq. \ref{PR} is used to designate the measurement between the platform and satellite $j$.

  \begin{equation}
   \rho^{j}_{IF} = \rho^{j}_{L1}\left[ \frac{f^{2}_{1}}{f^{2}_{1} - f^{2}_{2} } \right] - \rho^{j}_{L2}\left[ \frac{f^{2}_{2}}{f^{2}_{1} - f^{2}_{2} } \right]
   \label{PR}
  \end{equation}

  Using the IF combination, the pseudorange measurements are modeled as shown in Eq. \ref{prErr}: where $\delta t_{u}$ is the receiver's clock bias, $\delta t_{s}$ is the GNSS satellite's clock bias, $T^{}_{z,d}$ is the tropospheric delay in the zenith direction, and $\mathcal{M}_d(el^{j})$ is a user to satellite elevation angle dependent mapping function, $\delta_{Rel.}$ is the relativistic correction, $\delta_{P.C.}$ is the phase center offset between the orbit products and the propagating satellites antenna, and $\delta_{D.C.B}$ is the correction term to mitigate the differential code bias. For this study, all of the above mentioned models were are incorporated through the utilization of GPSTk~\cite{gpsTK}, which is an open-source GNSS software package. All additional un-modeled error components are included in Eq. \ref{prErr} as the $\epsilon$ term

  \begin{equation}
   \begin{split}
    \rho^{j}_{IF}  = & R_j + c(\delta t^{}_{u} - \delta t_{s})+ T_{z,d}  \mathcal{M}_d(el^{j}) \\
    & + \delta_{Rel.} + \delta_{P.C.} + \delta_{D.C.B} + \epsilon^{j}_\rho
    \label{prErr}
   \end{split}
  \end{equation}

  Using the modeled pseudorange observable, we can construct the pseudorange factor as shown in Eq. \ref{rhoFactor}. In Eq. \ref{rhoFactor}, the superscript $j$ designate the measurement between the platform and $j^{th}$ satellite, $\rho^j_{IF}$ is the observed pseudorange value, $\hat{\rho}^j_{IF}$ is the estimated pseudorange value as calculated in Eq. \ref{prErr}, and $\Sigma$ is the uncertainty in the pseudorange observable.

  \begin{equation}
   || e_{pr}^{j} ||_{\Sigma_{j}}^2 = || \rho^{j}_{IF} - \hat{\rho}^{j}_{IF} ||^2_{\Sigma_{j}} = e_{pr}^{j} \Sigma^{-1}_{j} e_{pr}^{j}
   \label{rhoFactor}
  \end{equation}

  \subsection{Fault Tolerance}

  Due to the degraded nature of the GNSS data in urban environments ( i.e., subject to multipath, poor satellite visibility ), methods to make the state estimation process more robust must be incorporated into the optimization routine. For this study, we will broadly classify these robust optimization techniques as traditional M-Estimators and more recent graph based techniques. Both methods are described in detail below.

  \subsubsection{Traditional M-Estimators}

  The field of M-Estimators dates back to the seminal paper published by Huber ~\cite{firstMEst}, which was later extended into a comprehensive survey on the subject ~\cite{robustStats}.  The field of research related to M-estimation can be reduced to minimizing the influence of erroneous data by replacing the $L_2$ cost function with a modified cost function, $\rho(e_i)$, which penalizes observables that do not conform to the user defined observation model.

  All cost functions can be classified as being redescending or non-redescending. A redescending cost function conforms to the property that $ \lim_{e_i \to \infty} \psi(e_i) = 0, $ where $\psi(e_i) = \frac{d\rho(e_i)}{de_i}$, which implies that the weight approaches 0 as the magnitude of the residual approaches $\infty$. For this study, two M-estimators were selected, one M-estimator that is redescending --- the Cauchy cost function ~\cite{appliedStats} --- and one M-estimator that is non-redescending --- the Huber cost function ~\cite{firstMEst}.

  The m-estimator utilized in this study are shown in greater detail in  Table \ref{mEstimators}. In Table \ref{mEstimators}, $\rho(e_i)$ represents the modified cost function, $\psi(e_i)$ is the first derivative of the cost function with respect to the residual, $w(e_i)$ specifies the weight applied to corresponding entries in the information matrix, and k is the user defined M-estimator kernel width.

  \renewcommand{\arraystretch}{2.2}
  \begin{table}[htb!]
   \centering
   \caption{Selected M-Estimators }
   \resizebox{\columnwidth}{!}{
    \begin{tabular}{|| c  c c c ||}
     \hline
     M-Estimator & $\rho(e_i)$                             & $\psi(e_i)$               & $w(e_i)$                \\ [0.5ex]
     \hline\hline
     Huber $\left\{ \begin{array}{@{}ll@{}} \text{if} \ |x| \leq k \\ \text{if} \ |x| > k \end{array}\right.$
     & $\left\{ \begin{array}{@{}ll@{}} \frac{e_i^2}{2} \\ k(|e_i|-\frac{k}{2}) \end{array}\right.$  &
     $\left\{ \begin{array}{@{}ll@{}} e_i \\ ksign(e_i) \end{array}\right.$
     & $\left\{ \begin{array}{@{}ll@{}} 1 \\ \frac{k}{|e_i|} \end{array}\right.$ \\
     \hline
     Cauchy      & $\frac{k^2}{2}log(1+\frac{e_i^2}{k^2})$ & $\frac{e_i}{1+e_i^2/k^2}$ & $\frac{1}{1+e_i^2/k^2}$ \\ [1ex]
     \hline
    \end{tabular}%
   }
   \label{mEstimators}
  \end{table}

  \subsubsection{Graph Based Approaches}

  Next, three more recent advances in robust optimization: switchable constraints, dynamic covariance scaling, and max-mixtures are discussed in the context of robust GNSS optimization.

  \paragraph{Switch Constraints}

  Switchable Constraints were first introduced by ~\cite{switchConstraints} as a method to reject false positive loop-closure constraints in the simultaneous localization and mapping (SLAM) problem. Additionally, this method has been evaluated for its ability to mitigate the affect of multipath in urban environment ~\cite{sunderhauf2012multipath, sunderhauf2013switchable}. The robustness of this optimization technique is granted through the incorporation of a new switchable constraint for every factor of interest in the graph. The switchable constraint can be thought of as an observation weight that is be optimized concurrently with the state estimates. The modified cost function that must be optimized to find the minimum error over the entire graph is shown in Eq. \ref{sfCost}, where $s$ is the switchable constraint, $\gamma$ is the initial estimate for the switch constraint, and $\psi()$ is a real-valued function such that $\psi() \rightarrow \{ 1,0 \}$. A graphical depiction of the modified graph is shown in Figure 2, where the residual associated with measurement $m$ at epoch $n$ exceeds the residual threshold.

  \begin{equation}
  \begin{split}
  \hat{X}, \hat{S} = \argmin_{X,S} & \sum_i  \lvert \lvert e_{pr}^i \rvert \rvert^{2}_{\Sigma} \\ + & \sum_i  \lvert \lvert \psi(s_{i}) * e_{pr}^i \rvert \rvert^{2}_{\Sigma} \\ + &\sum_i  \lvert \lvert \gamma_i - s_i \rvert \rvert^{2}_{\Xi}
  \end{split}
   \label{sfCost}
  \end{equation}

  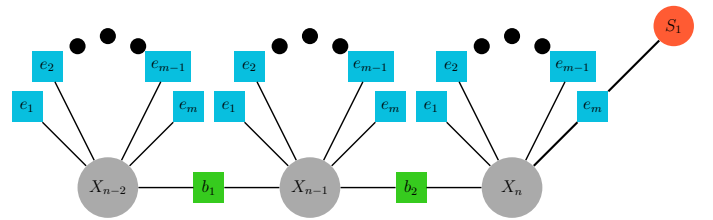
\begin{figure}[H]
   \begin{center}
    \begin{adjustbox}{width=0.5\textwidth}
     \begin{tikzpicture}
      \node[shape=circle,fill=grey, minimum size=1.5cm] (A) at (0,0) {$X_{n-2}$};
      \node[shape=circle,fill=grey, minimum size=1.5cm] (B) at (5,0) {$X_{n-1}$};
      \node[shape=circle,fill=grey, minimum size=1.5cm] (C) at (10,0) {$X_{n}$};

      \node[shape=rectangle,fill=blue, minimum size=0.75cm] (D) at (-2,2) {$e_{1}$};
      \node[shape=rectangle,fill=blue, minimum size=0.75cm] (E) at (-1.5,3) {$e_{2}$};
      \node[shape=circle,fill=black,minimum size=0.1cm] (a) at(-0.75,3.5) {};
      \node[shape=circle,fill=black,minimum size=0.1cm] (b) at(-0,3.75) {};
      \node[shape=circle,fill=black,minimum size=0.1cm] (c) at(0.75,3.5) {};
      \node[shape=rectangle,fill=blue, minimum size=0.75cm] (F) at (1.5,3) {$e_{m-1}$} ;
      \node[shape=rectangle,fill=blue, minimum size=0.75cm] (G) at (2,2) {$e_{m}$};

      \node[shape=rectangle,fill=green, minimum size=0.75cm] (P) at (2.5,0) {$b_{1}$};

      \node[shape=rectangle,fill=blue, minimum size=0.75cm] (H) at (3,2) {$e_{1}$};
      \node[shape=rectangle,fill=blue, minimum size=0.75cm] (I) at (3.5,3) {$e_{2}$};
      \node[shape=circle,fill=black,minimum size=0.1cm] (a) at(4.25,3.5) {};
      \node[shape=circle,fill=black,minimum size=0.1cm] (b) at(5,3.75) {};
      \node[shape=circle,fill=black,minimum size=0.1cm] (c) at(5.75,3.5) {};
      \node[shape=rectangle,fill=blue, minimum size=0.75cm] (J) at (6.5,3) {$e_{m-1}$} ;
      \node[shape=rectangle,fill=blue, minimum size=0.75cm] (K) at (7,2) {$e_{m}$};

      \node[shape=rectangle,fill=green, minimum size=0.75cm] (Q) at (7.5,0) {$b_2$};

      \node[shape=rectangle,fill=blue, minimum size=0.75cm] (L) at (8,2) {$e_{1}$};
      \node[shape=rectangle,fill=blue, minimum size=0.75cm] (M) at (8.5,3) {$e_{2}$};
      \node[shape=circle,fill=black,minimum size=0.1cm] (a) at(9.25,3.5) {};
      \node[shape=circle,fill=black,minimum size=0.1cm] (b) at(10,3.75) {};
      \node[shape=circle,fill=black,minimum size=0.1cm] (c) at(10.75,3.5) {};
      \node[shape=rectangle,fill=blue, minimum size=0.75cm] (N) at (11.5,3) {$e_{m-1}$} ;
      \node[shape=rectangle,fill=blue, minimum size=0.75cm] (O) at (12,2) {$e_{m}$};
      \node[shape=circle,fill=red, minimum size=1.0cm] (R) at (14,4) {$S_{1}$};

      \path [-,line width=1pt] (A) edge node {} (D);
      \path [-,line width=1pt] (A) edge node {} (E);
      \path [-,line width=1pt] (A) edge node {} (F);
      \path [-,line width=1pt] (A) edge node {} (G);

      \path [-,line width=1pt] (A) edge node {} (P);
      \path [-,line width=1pt] (P) edge node {} (B);

      \path [-,line width=1pt] (B) edge node {} (H);
      \path [-,line width=1pt] (B) edge node {} (I);
      \path [-,line width=1pt] (B) edge node {} (J);
      \path [-,line width=1pt] (B) edge node {} (K);

      \path [-,line width=1pt] (B) edge node {} (Q);
      \path [-,line width=1pt] (Q) edge node {} (C);

      \path [-,line width=1pt] (C) edge node {} (L);
      \path [-,line width=1pt] (C) edge node {} (M);
      \path [-,line width=1pt] (C) edge node {} (N);
      \path [-,line width=1.5pt] (C) edge node {} (O);
      \path [-,line width=1.5pt] (O) edge node {} (R);

     \end{tikzpicture}
    \end{adjustbox}
   \end{center}

   \caption{Example GNSS factor graph with switchable constraint}
  \end{figure}

  \paragraph{Dynamic Covariance Scaling}

  A pitfall of the switchable constraint robust optimization algorithm is that additional latent variable are added to the optimization process every time a observable's residual exceed the defined threshold. This  means that the optimizer is not only working over the original search space composed of the state vector, but is now also optimizing over all added switchable constraints. The inclusion of additional variables into the optimization process not only increases computation cost, but could also potentially decrease convergence speed. To overcome these issues, a closed formed solution to switchable constraints was introduced in ~\cite{DCS} as dynamic covariance scaling. The closed formed solution can be see in Eq. \ref{DCS}, where $\Phi$ is the inverse of the prior uncertainty on the switchable constraint, and $\chi$ is the initial error in the state.

  \begin{equation}
   s = min ( 1, \frac{2 \Phi}{ \Phi + \chi ^2} )
   \label{DCS}
  \end{equation}

  \paragraph{Max-Mixtures}

  The two previous robust models are still confined to uni-modal Gaussian error distributions, which in many situations will not fully capture the true distribution of the data in trying environments. To overcome this limitation, the next robust method extends that traditional uni-modal Gaussian factor graph optimization to a Gaussian mixture model. Generally, to move from a traditional uni-modal Gaussian optimization to a more complication distribution, the sum of multiple Gaussian components is utilized, as shown in Eq. \ref{sumGaus}.

  \begin{equation}
   p(z_i | x) = \Sigma_i w_i \mathcal{N}(\mu_i,\Lambda_i)
   \label{sumGaus}
  \end{equation}

  However, this solution greatly complicates the calculation of the maximum likelihood estimate because the logarithm cannot be ``pushed'' inside of the summation operator. To overcome this additional complexity, ~\cite{maxmix} recommend the utilization of the the max operator as a Gaussian component selector to approximation the true Gaussian summation. For GNSS data processing, this means that each observable can be modeled using two independent distributions: one distribution that defines data free of outliers, while a second distribution represents faulty data (i.e., the null hypothesis). The null hypothesis can be modeled as a Gaussian distribution centered at the mean of the error-free observable distribution, but with a substantially larger variance. This null hypothesis is tested against the hypothesis within the optimizer to iteratively find the weighting applied to the information matrix.

  \section{Algorithm Implementation And Performance}

  \subsection{Software Implementation}

  For implementation, this work uses the Georgia Tech Smoothing and Mapping Library (GTSAM)~\cite{dellaert2012factor} as the estimation engine. GTSAM is an open-source C++ library that abstracts the process of formulating an estimation problem as a factor graph. Important to our proposed application of robust GNSS, it allows developers to easily add custom measurement factor types, which allows for the flexible integration of different sensors. GTSAM also includes a variety of nonlinear optimizers for efficiently solving graphs over a specified subset of measurements.

  \subsection{Experimental Data Sets}

  To evaluate the algorithms positioning performance, two urban data-sets were collected -- the East-North ground trace for one of the collected data-sets is depicted in Figure \ref{enTrace} --  that contain multiple scenarios which are known to degrade GNSS data. A depiction of the encountered scenarios are provided in Figure 3. To see how these adverse scenarios can affect GNSS positioning performance, the real-time, tightly-coupled GNSS/INS NovAtel positioning solution is compared to the post-processed RTK solution in Figure \ref{novatelError}. As can be seen in Figure \ref{novatelError}, the adverse environments when driving in an urban setting can induce large fluctuations in positioning error.

  These data-sets provide approximately 4 hours of GNSS/INS data. Each  data-set contains dual-frequency GNSS pseudorange and carrier-phase observables at 10 Hz. Additionally, 50 Hz IMU data measured by the navigation grade IMU contained in the NovAtel SPAN system is included.

  \begin{figure}[H]
   \begin{center}
    \begin{subfigure}[b]{0.3\textwidth}
     \includegraphics[width=\textwidth]{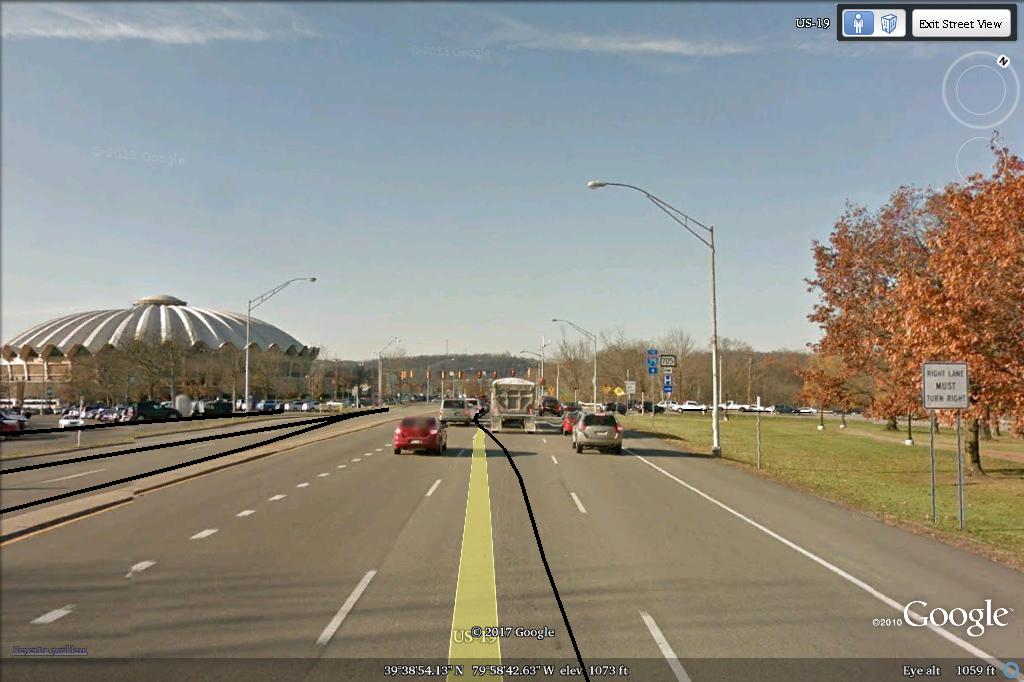}
     \caption{Clear sky}
    \end{subfigure}
    \begin{subfigure}[b]{0.3\textwidth}
     \includegraphics[width=\textwidth]{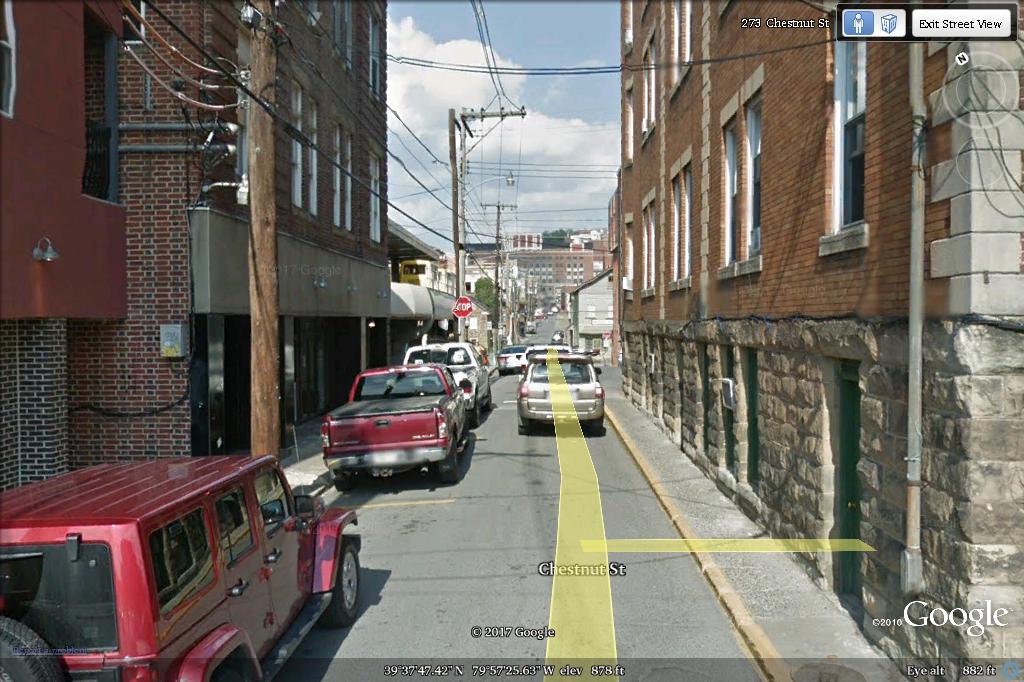}
     \caption{Partial satellite occlusion}
    \end{subfigure}
    \begin{subfigure}[b]{0.3\textwidth}
     \includegraphics[width=\textwidth]{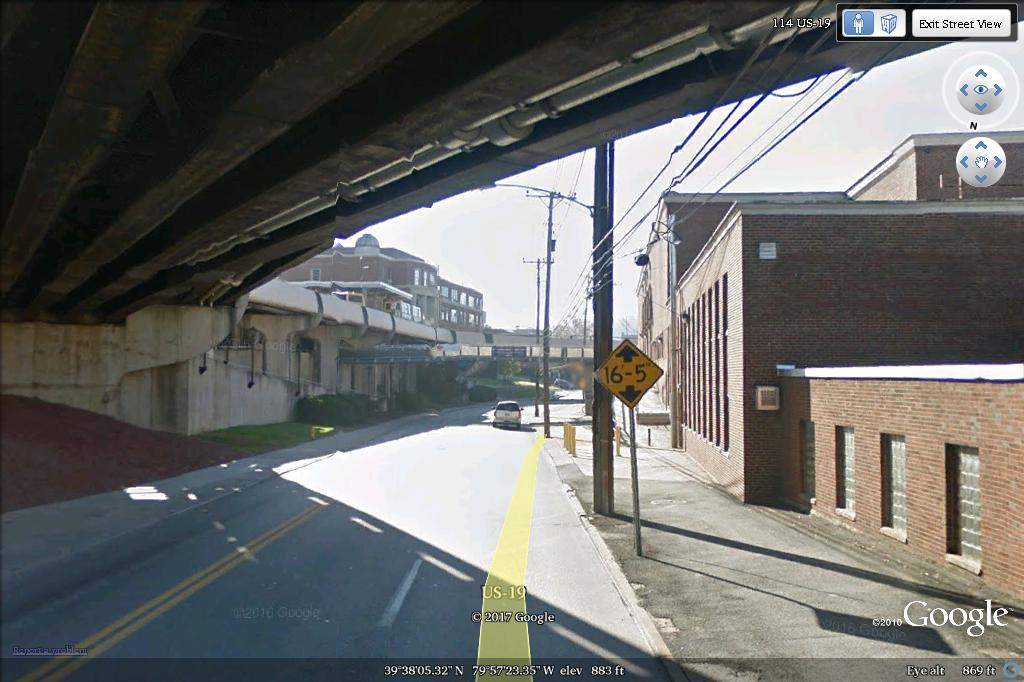}
     \caption{Total satellite occlusion}
    \end{subfigure}
    \label{dataEnv}
    \caption{Multiple scenarios encountered during the GNSS data collect. Two of which are known to degrade GNSS observables (i.e., partial and total satellite occlusion).}
   \end{center}
  \end{figure}

  \begin{figure}[htb!]
   \centering
   \includegraphics[width=0.5\textwidth]{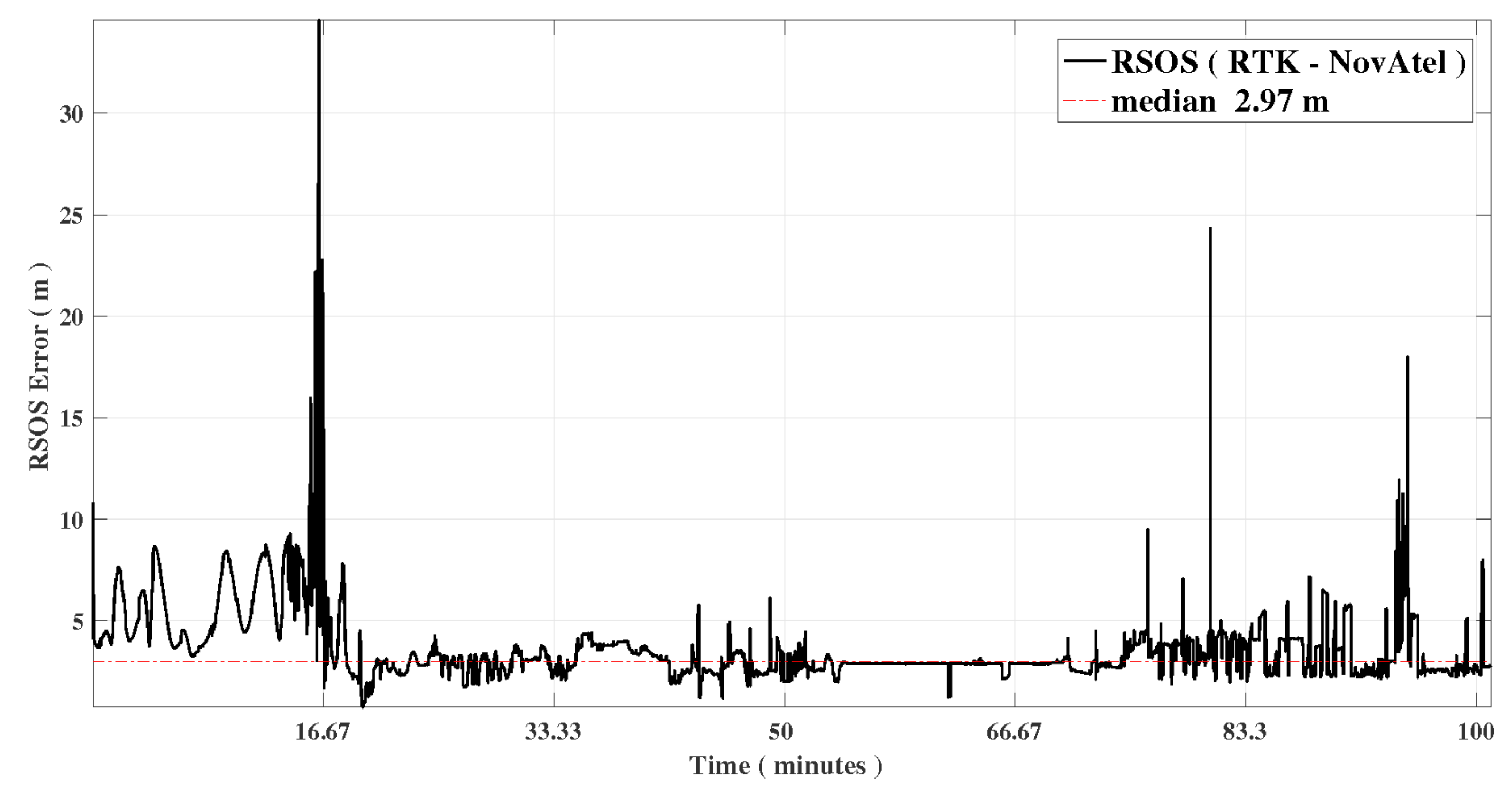}
   \caption{RSOS positioning difference between the reference solution generated by RTK and the positioning solution generated by tightly-coupled GNSS/INS NovAtel's real-time filter.}
   \label{novatelError}
  \end{figure}

  \begin{figure*}[htb!]
   \centering
   \includegraphics[width=\textwidth]{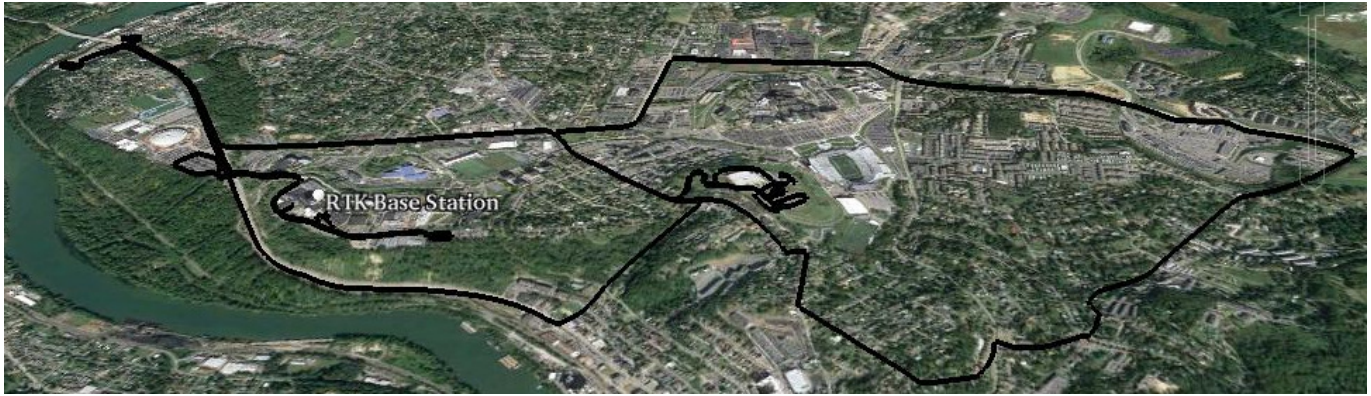}
   \caption{East-North ground trace for example data-set generated by Google Earth.}
   \label{enTrace}
  \end{figure*}

  \subsection{Reference Solution}

  To generate an accurate comparison solution, static data was collected concurrently with the roving platform. The static data sets provide 1 Hz dual-frequency GNSS pseudorange and carrier-phase observables collected by a NovAtel SPAN receiver. This data is used to generate a Kalman filter/smoother Carrier-Phase Differential GPS (CP-DGPS) positioning solution with respect to the local reference station. The CP-DGPS reference solution was calculated using RTKLIB \cite{rtklib}.

  \subsection{Evaluation}

\begin{figure}[htb!]
   \centering
   \includegraphics[width=0.5\textwidth]{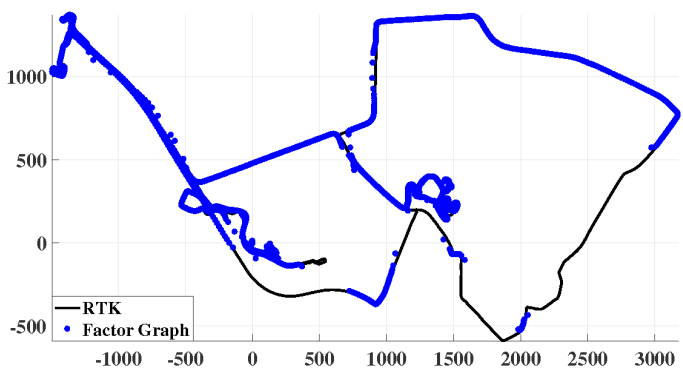}
   \caption{East-North positioning comparison in meters between the reference solution and the factor graph.}
   \label{compEN}
  \end{figure}

  Using  the data-set shown in Figure \ref{enTrace}, the factor graph formulation with the above mentioned noise models can be tested against the RTK positioning solution. As a visual comparison, the RTK positioning solution is compared to the solution obtained when utilizing a factor graph with $L_2$ optimization in Figure \ref{compEN}. As seen in the figure, the two solutions agree agree fairly well.

  The results for the six cost functions detailed in the previous sections is provided in Table \ref{noFaultRes}. This Table provides the mean, median, and max deviation of Residual Sum of Squares (RSOS) difference between the RTK positioning solution and the factor graph solution. From this result it should be noted that no one optimization routine has a clear advantage. This maybe attributed to the fact that --- although the data was collected in an environment that will degrade the observables (i.e., building induced multipath, and poor satellite geometry) --- the environment may not have been harsh enough to fully leverage the benefits that can be gained by the robust noise models.

  \begin{table}[htb!]
   \centering
   \caption{Optimization statistic comparison when no artificial faults are included.}
   \begin{tabular}{|| c  c c c ||}
    \hline
                  & median (m) & mean (m) & max (m) \\ [0.5ex]
    \hline\hline
    $L_2$         & 4.27       & 10.58    & 427.38  \\
    \hline
    Huber         & 5.16       & 12.30    & 450.00  \\
    \hline
    Cauchy        & 5.17       & 12.50    & 873.15  \\
    \hline
    DCS           & 4.28       & 10.63    & 547.42  \\
    \hline
    Switch Factor & 4.29       & 11.07    & 564.44  \\
    \hline
    Max-Mix       & 4.29       & 11.34    & 566.50  \\ [1ex]
    \hline
   \end{tabular}
   \label{noFaultRes}
  \end{table}

  To more clearly understand when the robust noise models become beneficial, simulated faults were incorporated into the study. The artificial faults were simulated as Gaussian random variable with a mean of zero and a standard deviation of 50 meters. The simulation faults were add onto the measured pseudorange observable at random, where it is possible that up to 49\% of the observables can contain artificial faults.

\begin{figure}[htb!]
   \centering
   \includegraphics[width=0.5\textwidth]{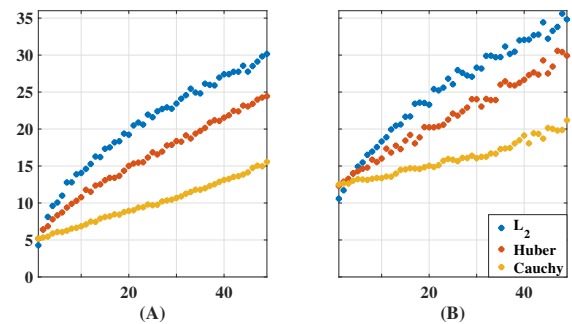}
   \caption{RSOS positioning error in meters for M-Estimators as the percentage of artificially degraded observations is increased (A) depicts the median RSOS positioning error and (B) depicts the mean RSOS positioning error.}
   \label{mEstComp}
  \end{figure}

  With the artificial faults added to the pseudorange observables, we will first look at how the traditional M-Estimators, which are detailed in Table \ref{mEstimators}, perform. The RSOS positioning error increase as the percentage of faults increases is depicted in Figure \ref{mEstComp}. From Figure \ref{mEstComp}, a clear advantage can be seen with the M-Estimators with respect to both the mean and median RSOS positioning error when compared to traditional $L_2$ optimization. When a comparison is conducted only between the M-Estimators, a rather large RSOS positioning error decrease is granted when the Cauchy cost function is utilized in place of the Huber. This benefit is not believed to be a general statement about the estimators. Instead, it is probably a byproduct of the limited diversity of data utilized for this study. For example, it was noted in ~\cite{emRobust} --- where they were analyzing robust loop closure methods for solving the SLAM problem --- that both the Cauchy and the Huber performed similarly.

\begin{figure}[htb!]
   \centering
   \includegraphics[width=0.5\textwidth]{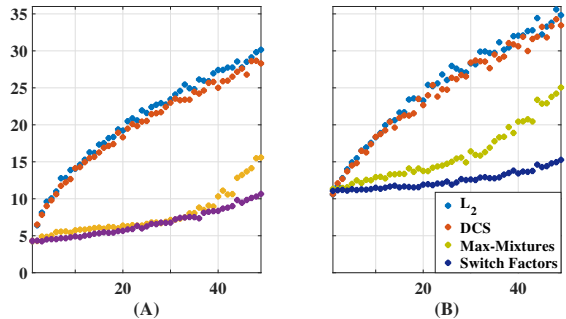}
   \caption{RSOS positioning error in meters for graph based robust methods as the percentage of artificially degraded observations is increased (A) depicts the median RSOS positioning error and (B) depicts the mean RSOS positioning error.}
   \label{graphComp}
  \end{figure}

  The same analysis was conducted for the graph based robust inference methods. The RSOS positioning error as a function of the percentage of artificial faults can be seen in Figure \ref{graphComp}. From this plot it can be noted that the dynamic covariance scaling is performing considerably worse than any of the other robust estimation techniques evaluated in this study. Additionally, it should be noted how relatively little the positioning performance is affected when switchable constraints or the max-mixtures approaches are utilized at lower observation fault probabilities; however, when the data is highly faulty (i.e., when around 35 percent of the observations contain faults) the switch factor method provides a clear advantage.

  \section{Conclusion}

  For autonomous navigation systems, there is a requirement that the system can robustly infer its desired states when confronted with adverse environments. To meet this requirement, we evaluated several robust optimization techniques from various fields for the application of GNSS pseudorange data processing in an urban environment. From this analysis, it has been shown that traditional M-Estimators can aid the optimization scheme when confronted with an adverse environment; however, it has also been shown that including Switchable Constraints in the graph considerably outperforms the other methods implemented in this study, specifically as the number of faulty observables increases.

  Additionally, to allow the GNSS community to build upon the advances made within other research communities, the software developed and data sets used for this evaluation has been released publicly at \url{https://github.com/wvu-navLab/RobustGNSS}, which provides a GNSS front-end integrated with several robust noise models to the GTSAM library.

  \section*{Acknowledgment}

  The material in this report was approved for public release on the $26^{th}$ of June 2017, case number 88ABW-2017-3109.

  \bibliographystyle{ieeetr}
  \bibliography{ionPaperSources}

\end{document}